\documentclass[11pt,a4paper]{article}
\usepackage[hyperref]{acl2019}
\usepackage{times}
\usepackage{latexsym}
\usepackage{enumitem}
\usepackage{times}  
\usepackage{helvet}  
\usepackage{url}
\usepackage{courier}  
\usepackage{url}  
\usepackage{graphicx}  
\usepackage{lipsum}

\usepackage{amsmath}
\usepackage{amsfonts,amssymb}
\usepackage{caption}
\usepackage{graphicx}
\usepackage{float}
\usepackage[utf8]{inputenc}
\usepackage{booktabs}
\usepackage{color}
\usepackage{amssymb}
\usepackage{url}
\usepackage{xcolor}
\usepackage{soul}
\usepackage{amsmath}
\usepackage{soulpos}
\usepackage{enumitem}
\usepackage[font={small}]{caption}
\usepackage{multirow}
\usepackage{graphicx}
\usepackage{mathtools}

\colorlet{nured}{red!40}
\colorlet{nublu}{blue!35}
\colorlet{nugre}{green!40}

\newcommand{\hlred}[1]{{\sethlcolor{nured}\hl{#1}}}

\newcommand{\hlblue}[1]{{\sethlcolor{nublu}\hl{#1}}}
\ulposdef{\ulnumaux}{%
   $\underset{\saveulnum}{\rule[-.7ex]{\ulwidth}{.4pt}}$}

\aclfinalcopy 

\title{Joint Entity Extraction and Assertion Detection for Clinical Text}

\author{  Parminder Bhatia \\ Amazon, USA \\  {parmib@amazon.com} \And
          Busra Celikkaya \\ Amazon, USA \\ {busrac@amazon.com } \And
           Mohammed Khalilia \\ Amazon, USA \\  {khallia@amazon.com}}

\date{}

\begin{document}
\maketitle

\begin{abstract}
Negative medical findings are prevalent in clinical reports, yet discriminating them from positive findings remains a challenging task for information extraction. Most of the existing systems treat this task as a pipeline of two separate tasks, i.e., named entity recognition (NER) and rule-based negation detection. We consider this as a multi-task problem and present a novel end-to-end neural model to jointly extract entities and negations. We extend a standard hierarchical encoder-decoder NER model and first adopt a shared encoder followed by separate decoders for the two tasks. This architecture performs considerably better than the previous rule-based and machine learning-based systems.  To overcome the problem of increased parameter size especially for low-resource settings, we propose the \textit{Conditional Softmax Shared Decoder} architecture which achieves state-of-art results for NER and negation detection on the 2010 i2b2/VA challenge dataset and a proprietary de-identified clinical dataset.

\end{abstract}
\section{Introduction}
In recent years, natural language processing (NLP) techniques have demonstrated increasing effectiveness in clinical text mining. Electronic health record (EHR) narratives, e.g., discharge summaries and progress notes contain a wealth of
medically relevant information such as diagnosis information and adverse drug events. Automatic extraction of such information and representation of clinical knowledge in standardized formats \cite{tomar2019relation} could be employed for a variety of purposes such as clinical event surveillance, decision support \cite{jin2018improving}, pharmacovigilance, and drug efficacy studies.

\begin{figure}[h]
\centering
\begin{tabular}{l}
\toprule
\hlblue{Discontinue} \hlred{Abraxane}, patient \hlblue{denies} taking\\ \hlred{Tyleno 325 mg} and is \hlblue{not} taking \hlred{calcium}\\ \hlred{carbonate}.  Patient also \hlblue{stopped} taking\\ \hlred{colecalciferol 1,000 units PO}.\\
\bottomrule
\end{tabular}
\caption{Negated medications (highlighted in red) and negation cues (highlighted in purple) in clinical text. Our model does not explicitly label the cues.}
\label{figure:annot}
\end{figure}
 
Although many NLP applications that successfully extract findings from medical reports have been developed in recent years, identifying assertions such as positive (present), negative (absent), and hypothetical remains a challenging task, especially to generalize \cite{wu2014negation}. However, identifying assertions is critical since negative and uncertain findings are frequent in clinical notes (Figure \ref{figure:annot}), and information extraction algorithms that do not distinguish between them will not paint a clear picture of the patient. 

In this paper, we focus on identifying the negated findings in a multi-task setting  \cite{bhatia2018dynamic, comprehendmedical}. Most of the existing systems treat this task as a pipeline of two separate tasks, i.e., named entity recognition (NER) and negation detection. Previous efforts in this area include both rule-based and machine-learning approaches. 

Rule-based systems rely on negation keywords and rules to determine the cue of negation. NegEx \cite{chapman2001simple} is a widely used algorithm that consists of ontology lookup to index findings, and negation regular expression search in a fixed scope. ConText \cite{harkema2009context} extends NegEx to other attributes like hypothetical and make scope variable by searching for a termination term. NegBio \cite{peng2018negbio} uses a universal dependency graph for scope detection. Another similar work by Gkotsis et al. \shortcite{gkotsis2016don} utilizes a constituency-based parse tree to prune out the parts outside the scope. However, these approaches use rules and regular expressions for cue detection which rely solely on surface text and thus are limited when attempting to capture complex syntactic constructions such as long noun phrases.

Kernel-based approaches are also very common, especially in the 2010 i2b2/VA task of predicting assertions. The state-of-the-art in that challenge applies support vector machines (SVM) to assertion prediction as a separate step after concept extraction \cite{de2011machine}. They train classifiers to predict assertions of each concept word, and a separate classifier to predict the assertion of the whole concept. Shivade et al. \shortcite{shivade2015extending} propose an Augmented Bag of Words Kernel (ABoW), which generates features based on NegEx rules along with bag-of-words features. Cheng et al. \shortcite{cheng2017automatic} use CRF for classification of cues and scope detection. These machine learning based approaches often suffer in generalizability, the ability to perform well on unseen text.

Recently, neural network models by Fancellu et al. \shortcite{fancellu2016neural} and Rumeng et al. \shortcite{rumeng2017hybrid} have been proposed. Most relevant to our work is that of Rumeng et al. \shortcite{rumeng2017hybrid} where gated recurrent units (GRU) are used to represent the clinical events and their context, along with an attention mechanism. Given a text annotated with events, it classifies the presence and period of the events. However, this approach is not end-to-end as it does not predict the events. Additionally, these models generally require large annotated corpus, which is necessary for good performance. Unfortunately, such clinical text data is not easily available.

 Multi-task learning (MTL) is one of the most effective solutions for knowledge transfer across tasks. 
 In the context of neural network architectures, we perform MTL by sharing parameters across models, such as pretraining using word embeddings \cite{bhatia2016morphological,bojanowski2016enriching}, a popular approach for most NLP tasks.
In this paper, we propose an MTL approach to negation detection that overcomes some of the limitations in the existing models such as data accessibility. MTL leverages overlapping representation across sub-tasks and it is one of the most effective solutions for knowledge transfer across tasks.
 In the context of neural network architectures, we perform MTL by sharing parameters across tasks.

To the best of our knowledge, this is the first work to jointly model named entity and negation in an end-to-end system. Our main contributions are summarized below:
\begin{itemize}[noitemsep,topsep=5pt]
    \item An  end-to-end hierarchical neural
model consisting of a shared encoder and different decoding schemes to jointly extract entities and negations. Using our proposed model, we obtain substantial improvement over prior models for both entities and negations on the 2010 i2b2/VA challenge task as well as a proprietary de-identified clinical note dataset for medical conditions.
    \item \textit{A Conditional softmax shared decoder} model to overcome low resource settings (datasets with limited amounts of training data), which achieves state of art results across different corpora.
    \item A thorough empirical analysis of parameter sharings for low resource setting highlighting the significance of the shared decoder.
\end{itemize}

\section{Methodology}
We first present a standard neural framework for named entity recognition.  
To facilitate multi-task learning, we expand on that architecture by building a two decoder model. Then, to overcome the issues of the two decoder model we propose a single shared decoder model. Finally, we introduce the \textit{Conditional softmax shared decoder}.

\subsection{Named Entity Recognition Architecture}\label{nerarch}
NER is a sequence tagging problem which maximizes a conditional probability of tags $\mathbf{y}$ given an input sequence $\mathbf{x}$, parameterized by $\theta$. 

\begin{equation}
    P(\mathbf{y} | \mathbf{x}; \theta) = {\displaystyle \prod_{t=1}^{T} P(y_t | x_t, y_{<t}; \theta)}
\end{equation}
Here $T$ is the length of the sequence, and $y_{<t}$ represents tags for all previous time-steps.
We focus on an established hierarchical architecture \cite{lample2016neural,yang2016multi,chiu2016named} consisting of encoders (at both word and character levels) and a tagger for output generation.

\subsubsection{Encoders} 
Input to the model, $\mathbf{x} \in \mathbb{N}^T$, represents token ids of the input vocabulary.
This sequence is encoded first at the character level and additionally at the word level.
Character level representation consists of using a bi-directional Long Short-Term Memory (BiLSTM) \cite{hochreiter1997long,graves2013speech} unit to encode each word independently.
For each word we subsequently have sequences $\overrightarrow{h^{(t)}_{1:l}}$ and $\overleftarrow{h^{(t)}_{1:l}}$, where $l$ represents the length of the word.
We concatenate the last time-step of each of these sequences to obtain a vector representation, $h_c^{(t)} = [\overrightarrow{h^{(t)}_l} || \overleftarrow{h^{(t)}_l}]$.
The final input to the word level encoder is a combination of a pre-trained word embedding \cite{pennington2014glove} and the character representation, $m_t = [h_c^{(t)} || \text{emb}_{word}(x_t)]$.
For the word level encoder we make use of another BiLSTM.


\subsubsection{Tagger} 
The tagger consists of a uni-directional LSTM which takes as input the latent word representation given by the word level encoder, as well as the label embedding of the previously generated tag.
During training we feed ground truth labels by way of teacher forcing \cite{williams1989learning}, while at test time we use the generated sequence directly.
This system is trained using a standard cross-entropy objective.

\begin{figure}[t!]
    \centering
    \includegraphics[scale=0.30]{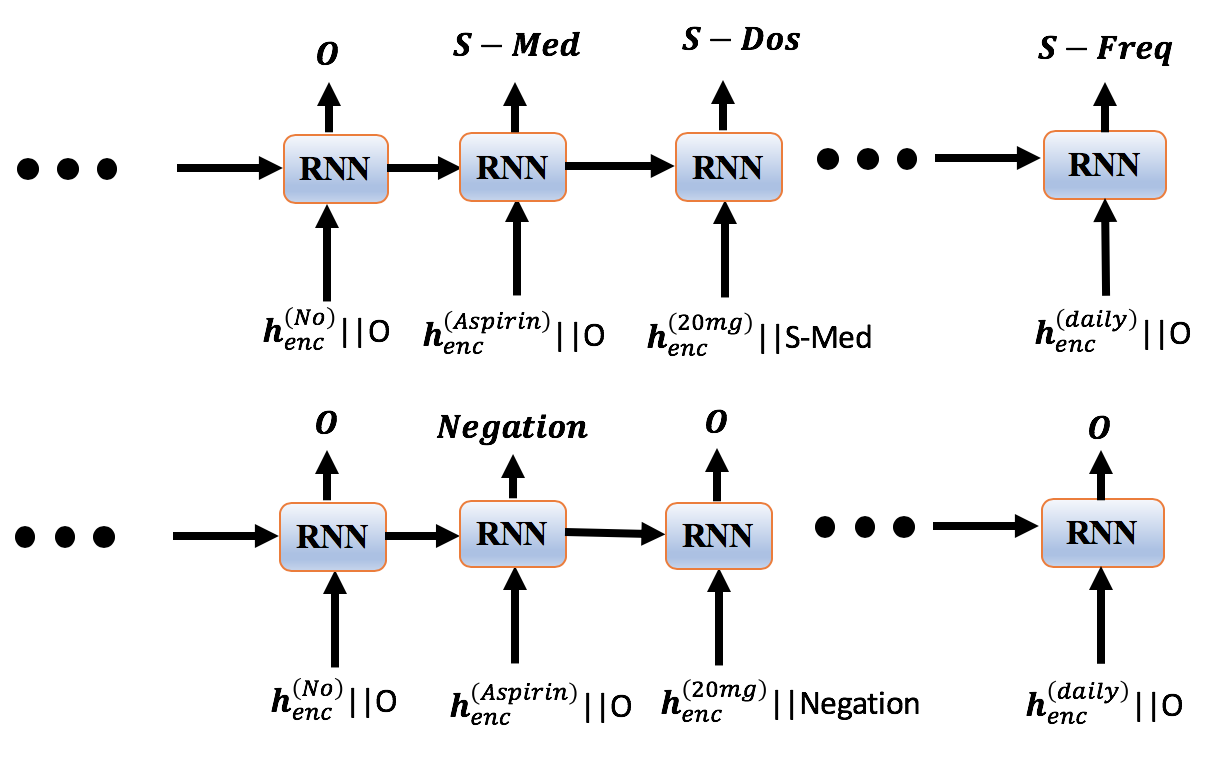}
    \caption{Two decoder model, upper decoder for NER and the lower decoder for negation, where common encoder}
    \label{fig:td}
    \label{figure:twodecmodel}
\end{figure}

\begin{table*}
\setlength{\tabcolsep}{6.7pt}
\renewcommand{\arraystretch}{1.05}
\centering
\begin{small}
    \begin{tabular}{lrrr|rrr}
        \toprule
        & \multicolumn{3}{c|}{2010 i2b2/VA} & \multicolumn{3}{c}{Proprietary Med. Cond.}\\
        \midrule
        Model & P & R & $\text{F}_1$ &  P & R & $\text{F}_1$\\ 
        \midrule
        \multirow{5}{*}{\rotatebox{90}{\textsc{NER}}} \quad LSTM:CRF \cite{chal2016ner}& 0.844 & 0.834 & 0.839  & 0.820 & 0.840 & 0.830\\
        \quad\quad Independent NER \cite{lample2016neural} & 0.857 & 0.841 & 0.848 & 0.880 & 0.848 & 0.863\\
        
        \quad\quad Two Decoder (this paper) & 0.849 & 0.855 & 0.851 & 0.876 & 0.861 & 0.868\\
        
        \quad\quad Shared Decoder (this paper) & 0.852 & 0.821 & 0.834 & 0.864 & 0.841 & 0.85\\
        
        \quad\quad \textbf{Conditional} (this paper) & 0.854 & 0.858 & \textbf{0.855} & 0.878 & 0.872 & \textbf{0.874}\\
        \midrule
        \multirow{6}{*}{\rotatebox{90}{\textsc{Negation}}} \quad Negex  \cite{chapman2001simple}&  0.896 & 0.799 & 0.845 & 0.403 & 0.932 & 0.563 \\
        \quad\quad ABoW Kernel \cite{shivade2015extending} & 0.899 & 0.900 & 0.900 & - & - & -\\
        \quad\quad Independent Negation \cite{lample2016neural} & 0.810 & 0.850 & 0.820 & 0.840 & 0.820 & 0.83\\
        
        \quad\quad Two Decoder (this paper) & 0.894 & 0.908 & 0.899 & 0.931 & 0.865 & 0.897\\
        
        \quad\quad Shared Decoder (this paper) & 0.870 & 0.902 & 0.882 & 0.921 & 0.850 & 0.878\\
        
        \quad\quad \textbf{Conditional} (this paper) & 0.919 &   0.891 & \textbf{0.905} & 0.928 &   0.874 & \textbf{0.899}\\
        
        \bottomrule
    \end{tabular}
    \end{small}
    \caption{Test set performance during multi-task training. The table displays precision, recall and macro averaged $\text{F}_1$.  The baseline is the current state-of-the art optimized architecture.}
\label{table:results}
\end{table*}

\subsection{Two Decoder Model}
To facilitate the MTL setting, we begin with a two decoder model consisting of decoders which use the shared encoder representation to jointly predict entities and negation attribute (Figure \ref{figure:twodecmodel}). This is a standard architecture used for MTL which consists of different LSTM's for decoders followed by corresponding softmaxes. This model mitigates the issues associated with rule-based models that rely solely on surface text, and thus are limited when attempting to capture complex syntactic constructions.
With shared contextual encoder representation consisting of character and word embedding based models, the proposed architecture provides an effective solution for knowledge transfer across tasks, thus consolidating the ability to perform well on unseen text.
However, this proposed architecture is not scalable, the number of decoders scales linearly with the number of attributes. Another problem we realized with this architecture is the performance degradation when working in an extremely low resource setting, where more parameters prevent the model from generalizing well.

\subsection{Shared Decoder Model}
\begin{figure}[t!]
     \centering
     \includegraphics[scale=0.3]{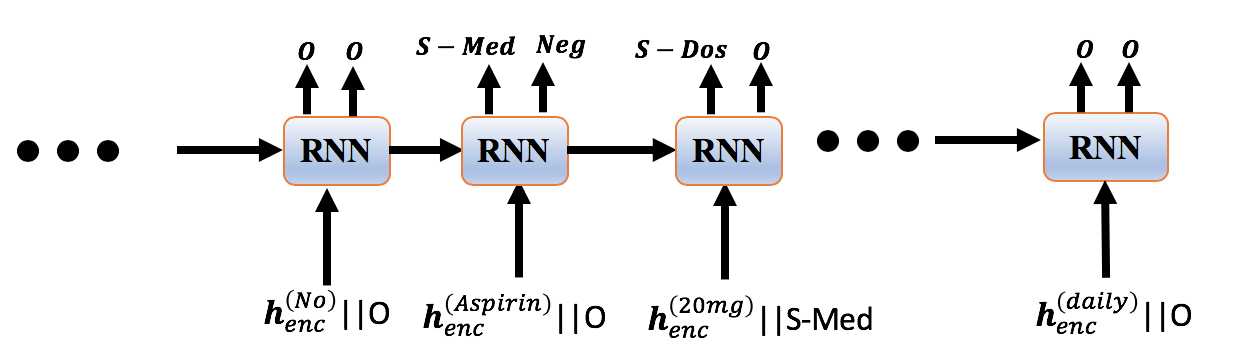}
     \caption{Shared decoder model}
    \label{figure:shareddecoder}
 \end{figure}

To overcome the limitations of the two decoder model we propose a shared decoder model (Figure \ref{figure:shareddecoder}). 
We share the encoder and decoder of the two tasks and the common output from the decoder is fed into two different softmax for entity and negations.
\begin{align*}
    \hat{y}^{Entity}_t &= \text{Softmax}^{Ent}(\mathbf{W^{Ent}}o_t + b^s)\\
    \hat{y}^{Neg}_t &= \text{Softmax}^{Neg}(\mathbf{W^{Neg}}o_t + b^s)
\end{align*}

\begin{figure}[h!]
    \centering
    \includegraphics[scale=0.4]{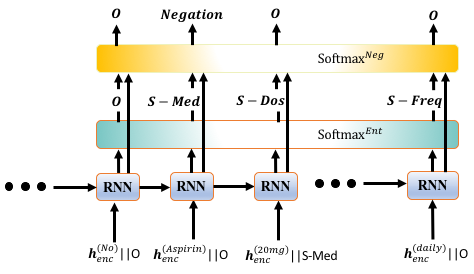}
    \caption{Conditional softmax decoder model}
    \label{figure:conditionaldecoder}
    
\end{figure}
\subsubsection{Conditional Softmax Decoder Model} 
While the single decoder model is more scalable, we found that this model did not perform as well for negation as the two decoder model. It can be attributed to the fact that negation occurs less frequently than the entities, thus the decoder primarily focuses on making entity extraction predictions. To mitigate this issue and provide more context to negation attributes, we add an additional input, which is the softmax output from entity extraction (Figure \ref{figure:conditionaldecoder}). Thus, the model learns more about the input as well as the label distribution from entity extraction prediction. As an example, we use negation only for \textsf{\scriptsize PROBLEM} entity in the i2b2 dataset. Providing the entity prediction distribution helps the negation model to make better predictions. The negation model learns that if the prediction probability is not inclined towards \textsf{\scriptsize PROBLEM}, then it should not predict negation irrespective of the word representation.

\begin{small}
\begin{align*}
    \hat{y}^{Ent}_t , \text{SoftOut}^{Ent}_t &=  \text{Softmax}^{Ent}(\mathbf{W^{Ent}}o_t+ b^s)\\
    \hat{y}^{Neg}_t &= \text{Softmax}^{Neg}(\mathbf{W^{Neg}}[o_t,\\
    &\phantom{{}=1}\text{SoftOut}^{Ent}_t] + b^s)
\end{align*}
\end{small}

where, $\text{SoftOut}^{Ent}_t$ is the softmax output of the entity at time step $t$. 

\section{Experiments}
\subsection{Dataset}
We evaluated our model on two datasets. First is the 2010 i2b2/VA challenge dataset for ``test, treatment, problem'' (TTP) entity extraction and assertion detection (\textit{i2b2 dataset}). Unfortunately, only part of this dataset was made public after the challenge, therefore we cannot directly compare with NegEx and ABoW results. We followed the original data split from R. Chalapathy and Piccardi \shortcite{chal2016ner} of 170 notes for training and 256 for testing. The second dataset is proprietary and consists of 4,200 de-identified, annotated clinical notes with medical conditions (\textit{proprietary dataset}).

        
   

\begin{figure}
    \centering
    \includegraphics[scale=0.4]{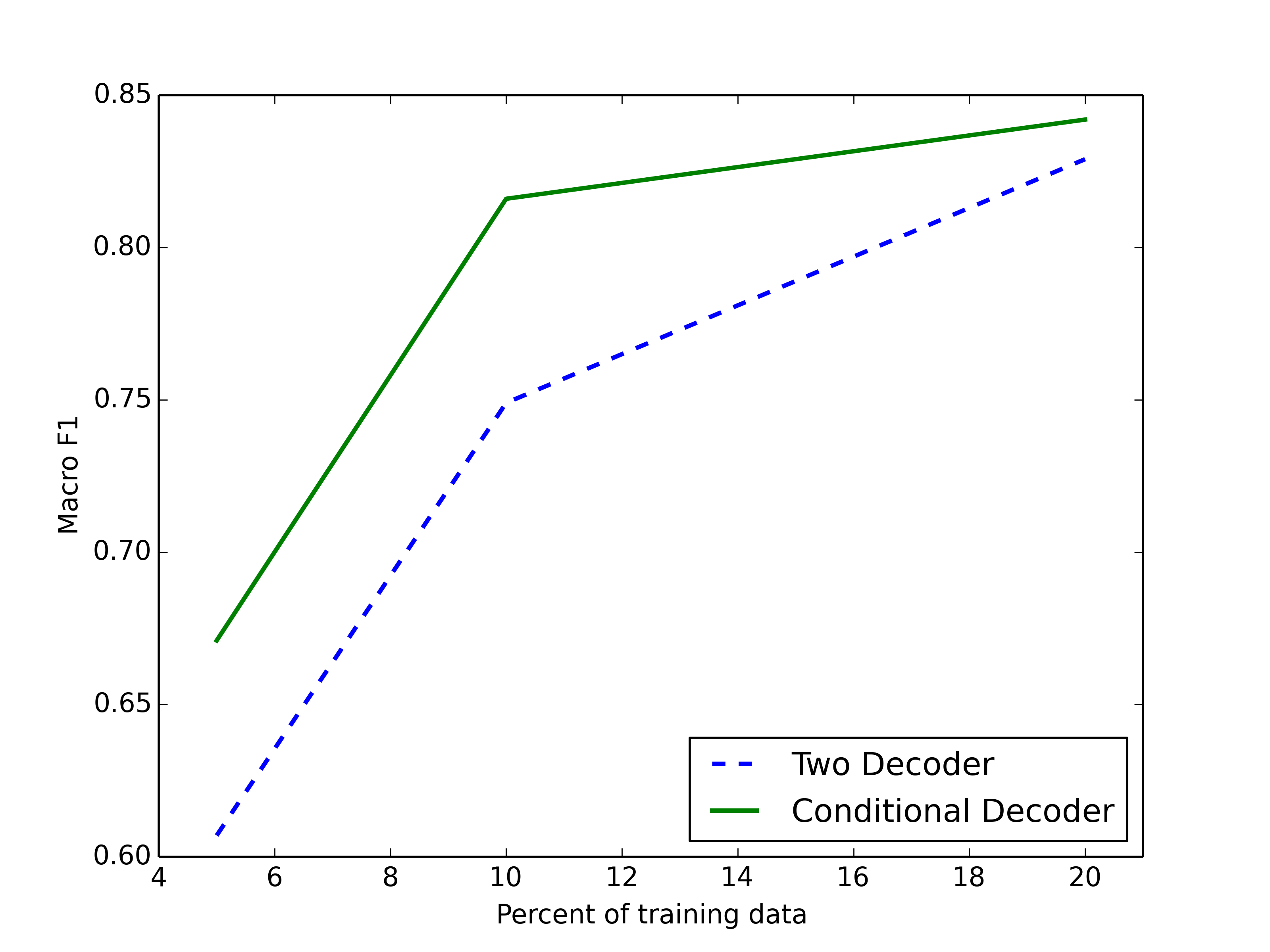}
    \centering
    \caption{Conditional softmax decoder is more robust in extreme low resource setting than its two decoder counterpart.}
    \label{fig:results_co}
\end{figure}

\subsection{Model settings}
Word, character and tag embeddings are 100, 25, and 50 dimensions, respectively.
For word embeddings we use GloVe \cite{peng2018negbio} and fine tune during training, while character and tag embeddings are randomly initialized.
Character and word encoders have 50, and 100 hidden units, respectively, while the tagger LSTM has a hidden size of 50.
Dropout is used after every RNN, as well as for word embedding input.
We use Adam \cite{kingma2014adam} as an optimizer.
Hyperparameters are tuned using Bayesian Optimization \cite{snoek2012practical}.


\section{Results}
Since there is no prior work which has solved the two tasks as a joint model, we report the best results for both the individual tasks (Table \ref{table:results}). We observe that the baseline model for NER (\textbf{Indepedent NER}) presented in the methodology section outperforms the best model \cite{chal2016ner} on the i2b2 challenge. The \textbf{Two decoder} and the conditional softmax decoder (\textbf{Conditional decoder}) model achieve even better results for NER than our baseline model, where the conditional decoder model achieved new state-of-art for 2010 i2b2/VA challenge task. \textbf{Shared decoder} underperformed the other two models. That can be attributed to a single decoder which primarily focuses on making entity extraction predictions which are more frequent than negations. 
The conditional decoder outperformed the baseline model on the negation prediction task and achieved an improvement of about $8\%$ in $\text{F}_1$ score compared to the baseline model, which suggests that modeling named entity and negation tasks together helps in achieving better results than each of the tasks done independently.

We compare our models for negation detection against NegEx, and ABoW which has best results for the negation detection task on i2b2 dataset. Conditional decoder model outperforms both NegEx and ABoW (Table \ref{table:results}). Low performance of NegEx and ABoW is mainly attributed to the fact that they use ontology lookup to index findings and negation regular expression search within a fixed scope. A similar trend was observed in the medication condition dataset. The important thing to note is the low $\text{F}_1$ score for NegEx. This can primarily be attributed to abbreviations and misspellings in clinical notes which can not be handled well by rule-based systems.

To understand the advantage of conditional decoder, we evaluated our model in extreme low data settings where we used a sample of our training data. We observed that the conditional decoder outperforms the two decoder model and achieved an improvement of $6\%$ in $\text{F}_1$ score in those settings (Figure \ref{fig:results_co}). As we increase the data size, their performance gap narrows in demonstrating that the conditional decoder is robust in low resource settings.

\section{Conclusion}

In this paper we have shown that named entity and negation assertion can be modeled in a multi-task setting. Joint learning with shared parameters provides better contextual representation and helps in alleviating problems associated with using neural networks for negation detection, thereby achieving better results than the rule-based systems. Our proposed conditional softmax decoder achieves best results across both tasks and is robust to work well in extreme low data settings. For future work, we plan to investigate the model on other related tasks such as relation extraction, normalization as well as the use of advanced conditional models.

\bibliography{acl2019}
\bibliographystyle{acl_natbib}

\appendix

\end{document}